# The Multi-phase spatial meta-heuristic algorithm for public health emergency transportation

Fariba Afrin Irany
University of North Texas
Denton
faribaafrinirany@my.unt.edu

Arnav Iyer
University of North Texas
Denton
ArnavIyer@my.unt.edu

Rubenia Borge Flores
University of North Texas
Denton
RubeniaBorgeFlores@my.unt.edu

Armin R. Mikler
Georgia State University
Atlanta
Armin.Mikler@unt.edu

**Abstract**

The delivery of Medical Countermeasures -MCMs for mass prophylaxis in the case of a bio-terrorist attack is an active research topic that has interested the research community over the past decades. The objective of this study is to design an efficient algorithm for the Receive Reload and Store Problem –RSS in which we aim to find feasible routes to deliver MCMs to a target population considering time, physical, and human resources and capacity limitations. For doing this, we adapt the p-median problem to the POD-based emergency response planning procedures and propose an efficient algorithm solution to perform the p-median in reasonable computational time. We present RE-PLAN, the Response PLan Analyzer system that contains some RSS solutions developed at The Center for Computational Epidemiology and Response Analysis (CeCERA) at the University of North Texas. Finally, we analyze a study case where we show how the computational performance of the algorithm can impact the process of decision making and emergency planning in the short and long terms.

**Introduction**

Points of Dispensing (PODs) are locations where people can collect medications, vaccines, or medical supplies during health emergencies. Each POD has a specific number of booths, and volunteers are associated with each booth to distribute medical countermeasures to people. The problem discussed in this paper is motivated by the POD based plan that is activated on an intentional or accidental release of biochemical substances. Federal and state governments decide to activate the POD bases response plan in an emergency, which results in a region's ability to provide life-saving medical countermeasures to its population.

Resources or medical countermeasures (MCMs) from the Strategical National Stockpile assets (SNS) will be delivered to state Receiving, Staging, and Storage (RSS) sites after the decision is made to activate the response plan. The medical countermeasures will be sent to the points of dispensing (PODs) from the RSS sites for distribution among people in need.

The delivery of medical countermeasures to PODs is of great concern for both planners and public health professionals. The distribution process from RSS sites to PODs requires trucks or other delivery vehicles for carrying medical countermeasures and a dedicated route for each vehicle. Several issues need to be considered while planning for delivery routes. First, we do not have unlimited time to distribute these assets to the PODs. In many cases, there is a specific time frame within which people need these medical countermeasures. After this period, the efficacy of vaccines reduces, and fatalities are expected to increase drastically. For example, people who received antibiotics within 4.7 days of exposure to anthrax had a 40% case fatality, but people who received treatment after this period had 75% case-fatality [1]. If we consider the total time available after the plan activation is 48 hours and the medical countermeasures (MCMs) will be delivered to the PODs within 12 hours, then the population allocated to each POD must be dispensed within 36 hours. If the delivery arrives late, each POD must distribute the same number of people with less time by serving more people in parallel by using more resources. The goal of the planners is to minimize the time to deliver the assets to the PODs. Planners must also consider the demand for each of the PODs. The combined order of all the PODs in a route must not exceed the truck capacity.

The purpose of developing tools to assist public health planners is to provide them with alternatives and resources that will contribute to their ultimate goal: to save and protect human life. One of the biggest challenges in public health is the distribution of medica countermeasures in the event of emergencies that demand mass distribution with significant limitations in-time and resources. Optimizing resources in a time of emergency is critical. In the context of natural disasters and terrorist attacks, it becomes a priority to develop solutions that can be used to save people's lives and minimize the negative impact.

The distribution of resources with considerable restrictions in-time, human resources, availability of supplies, and a massive number of people demanding a fast service is a complicated situation. It's required to look for solutions that will maximize the use of available resources, reduce waste, and allow practitioners to design plans of action that can adjust with real-world situations, problems, and solutions. It is necessary to look for automated solutions that take advantage of mathematical algorithms that can considerably reduce the time for finding optimal routes that satisfy the demands of the population within the timeframe that will allow serving the resources that will save human lives. Furthermore, a poorly designed plan could cause harm to people and communities, increase the destruction of material assets, and increase the negative impact of an already terrible disaster.

The Response Plan Analyzer (RE-PLAN) tool has been developed to assist public health planners in achieving their goals. RE-PLAN has been designed over the past ten years to emphasize the planning process over the creation of static plans to enable planners to quickly adapt to unfolding circumstances in the event of plan activation. It was designed to enable planners to build and analyze response plans designed to mitigate biological health threads or wide-spread medical emergencies. RE-PLAN enables planners to leverage geospatial population data to select Points of Dispensing (POD) locations, conducted needs-based vulnerability analysis on a per-POD basis, and estimate resource needs (personnel, Medical Counter Measures (MCM), etc.) at each POD. In addition to providing tools for MCM dispensing, RE-PLAN also includes a logistic module to enable planners to explore delivery options for the distribution of MCMs to PODs.

**Purpose of the Study**
The purpose of this research is to present the design and development of the multi-phase spatial meta-heuristic algorithm developed at the Center for Computational Epidemiology and Response Analysis (CeCERA) and to demonstrate its application in solving the problem generation of feasible routes required by public health planners to plan and execute mass prophylaxis. We develop a new algorithm, analyze its complexity, benchmark with other solutions proposed, and study its usability in the public health domain.

**Research problem investigated**
Our algorithm generates feasible routes for the RSS problem that is given a specific area and its corresponding population; provide feasible routes that comply with the current time and capacity constraints and the particularities of mass prophylaxis in the public health domain. Currently, there is not a proper tool available for public health planners that will aid them in generating plans for medical countermeasures distributions in the event of public health emergencies. This represents a severe problem because planning for routes generation to distribute resources in large populations is complicated, time-consuming, and unfeasible if the planner does not have access to accurate information. We propose the application of the multi-phase meta-heuristic algorithm to solve this problem because the consequences of not having available system tools can lead to weak and disastrous response for resource distribution in the event of public health emergencies in medium and large scale.

**Time sensitivity**
The distribution of medicine and supplies is highly time-sensitive, more so in cases of significant and severe incidents. For instance, in an attack with inhaled anthrax, the victims need to receive antibiotics within 48 hours of infection because, after that, the probability of survival decreases significantly. With the threats of nuclear weapons, bioterrorism, and even natural disasters due to climate changes, it becomes imperative to develop an emergency response plan that will serve the community in the minimum time possible. Time is critical as a response that is time-efficient will help to save human lives, prevent and stop disease spread, protect communities that haven't been affected, and even help to reduce material damage.

**Importance of minimizing routes**
One of the challenges in public health departments and other departments in different areas is the limitation and sometimes scarcity of resources. In the context of public emergencies, it is essential to reduce waste at the same time that we generate feasible solutions. Our algorithm increases efficiency in the planning process by minimizing the number of routes required to serve a population. The multi-phase spatial meta-heuristic algorithm reduces the amount of the routes needed to execute a plan by loading each vehicle to its maximum capacity before assigning a route. Minimizing the number of routes in a delivery plan reduces the waste in resources: vehicles, transportations costs, delivery time, drivers, and security personnel to guard the vehicles.

**Literature review**

A time-space network model was developed to deliver relief materials to the people in need in a public health emergency. In this model, supplies and demand varied with time, and more than one mode was used for delivery. The authors in this paper decomposed the model into two network flow problems, each having multiple periods and multiple commodities [2]. A description of recent improvements of advancement models and commitments, with an accentuation on their applications in circumstances of vulnerabilities and inconsistency, as realistic in low-asset nations, is introduced [3]. And for solving this problem, they used a generic algorithm [4]. A heuristic called adaptive LNS was proposed to solve the problem of delivery using the few vehicles available and within the time window. The authors got a satisfactory result when they tested in real life [5]. A column generation-based algorithm was proposed where a heuristic method was used to solve the problem of the vehicle routing problem, having demand that depends on the time [6]. The epidemic diffusion rule was used to build a changing optimization model with demands varying with time for emergency logistics [7]. A heuristic algorithm was developed for ambulance redeployment, and the goal was to minimize the number of late arrivals. The authors of the paper evaluated the performance of the heuristic using a simulated model of an emergency [8]. A Tabu Search heuristic (TS) was proposed for transferring the wounded people in a large-scale public health emergency [9]. Adaptive Large Neighborhood Search was used to solve the inventory slack routing problem (ISRP). The problem is different from any other vehicle routing problem as it does not use the cost objective function. The inventory slack routing problem (ISRP) aims to augment slack in delivering stuff to the site, and the sites can operate without any problem if there is any interruption [10]. The authors of the paper developed a route optimization model where the number of delivery vehicles required is less the demand areas, and they used Fish-Swarm Ant Colony Optimization (FSACO) for finding a solution for the model [11]. The environmental deterioration was given priority and merged with the optimization concept of emergency delivery, and thus a fuzzy low-carbon open location-routing problem (FLCOLRP) was proposed. This paper created consciousness among the people about the emergency that can arise in the environment after a calamity or intentional attack [12]. A Capacitated Vehicle Routing Problem was proposed, which can synchronize according to the activity, considering the cases where the cost of transporting the medication is higher than the cost of medicine production. There proposed solution took into account the cases when enough medicines are not available in a hospital and also the cases when an excess amount of medications are ordered without necessity and causes financial losses [13]. This paper concerns the plan of supplying relief material using satellite distribution centers (SDCs) in case of emergencies. It presents the idea of split delivery, which can be solved by a heuristic approach proposed by the authors [14]. An integrated mathematical model was developed considering the flow to relief material, mixed schedules of pickup and distribution, and a facility location problem with four layers [15]. A heuristic algorithm called the Emergency Relief Transportation Planning Algorithm (ERTPA) capable of sorting demands basing on several factors such as due dates, distances of the demand points from the delivery point was proposed. The authors also developed a model to test the efficacy of their proposed solution [16]. Natural disasters like a tornado can make the roots of the tree toppled, rip households to smaller parts, spread the garbages over the road due to the high speed of the wind, and result in the blockage of the road. This type of situation obstructs the transfer of relief materials to the Distribution Centre within the limit. There is not much work done until now, which focuses on the uncertainties associated with the travel time of routing vehicles during an emergency like natural disasters or human-made attack. The authors proposed a new model named ACO-TS by combining the ant colony optimization (ACO) and tabu search (TS) to solve the multiobjective rescue routing (MORR) model established by them considering the situations which focus on uncertainty in travel time [17]. For reducing the effort needed to solve the problem of relief material delivery as an integer program, the authors of the paper divided the problem into two subproblems. One is the routing problem, which deals with the distribution of the medications among the PODs, and the other is the determination of the start time of each vehicle dedicated to a route and demand associated with each POD [18]. The authors of the paper tested the two-stage approach using one-county and three-county scenarios for the state of Maryland, but they did not describe the test elaborately in the paper. In an emergency, transferring relief materials to the people within the time limit may not be sufficient as there can be cases when people are injured. In this case, transferring the injured people is an indispensable part of the relief operations. To provide maximum support using minimum cost, a forward and backward relief network capable of accomplishing multiple objectives was proposed to help injured people transfer to the hospitals as well as provide medical supplies to the people after a catastrophe [19]. This network uses two phases of disaster where the post-disaster phase deals with two routes one in the forward direction and another in the backward direction, each one performing a single operation. Natural disasters like an earthquake can be devastating if proper necessary post-earthquake steps are not taken. For instance, people injured in an earthquake need to be transferred immediately from the earthquake-affected area to a safe place where they will get proper medical aids. Along with shelter and medical help, food supply needs to be ensured for the people in the post-earthquake phase. Otherwise, people will suffer from hunger, affected with the disease, and eventually can die. Keeping the shelters in a store as a pre-earthquake measure and distribution of the shelters to the short-term shelter area in the post-earthquake phase is an important step in mitigating the damage of lives associated with an earthquake. A multi-purpose mixed-integer linear programming model was developed to help decision-makers with the storage and distribution of earthquake shelters [20]. The time of an earthquake attack, the damage caused by it, and the amount of relief materials needed at the post-earthquake phase is uncertain. For minimizing the time and cost associated with the post-earthquake relief material materials distribution, a multi-purpose model for the fuzzy location-routing problem (LRP) was set up, and a genetic algorithm based on weighted coefficient transformation capable of solving the vehicle routing problem was put forward [21]. It is not feasible to deliver the relief material to every single home after a disaster. It is better to choose some easily reachable points as a distribution center from where people can collect relief materials. The authors of the paper chose Satellite distribution Centers as the point of distributions as these are within walking distance from home. But the main challenge is to deliver the relief materials from the central depot to this distribution centers using a fleet of vehicles, each having a certain capacity. A covering tour approach was used by the authors to model this situation, and they developed a heuristic method for solving this problem[14]. The local search heuristics proposed that consider using spatial neighborhood in Voronoi diagrams for developing local search is not much. The authors of the paper focused on solving wide-ranging distance constrained capacitated vehicle-routing problems by using Voronoi spatial neighborhood-based search heuristic and algorithm, which was evaluated using four sets of benchmark tests for 200–8683 customers and gave better solution compared to many methods developed previously [22]. The authors of the paper gave the research gaps associated with emergency logistics priority. They came up with a framework capable of filling these gaps by prioritizing three essential things: location of the short-term unit for an emergency, the method of transferring injured people to the emergency-unit, and the number of people assigned for each unit [23]. The two critical parts of emergency logistics are to locate the distribution center from where people can collect the relief materials and the vehicle routing problem, which involves the transfer of these relief materials from the central depot to these distribution centers. In most cases, these two elements are considered as an independent problem and solved separately. The authors of the paper felt the necessity of solving these two problems simultaneously and came up with a model aiming to minimize the wait time of the recipients [24]. There are many cases when transferring the medical aids to the distribution center within the time limit is not possible due to the long-distance, location of the distribution center in an isolated place or damage caused in the road due to disaster. A two-stage approach was proposed by the authors of the paper to solve the problem of emergency logistics in such cases. The first stage proposed a fuzzy method and its heuristic to choose the temporary distribution centers and medical aid points assigned to each interim distribution Center. In the second stage, an integer-programming model was proposed for solving the problem of delivery route determination [25].

**Algorithm:**
The tool REPLAN was designed in such a way that it can use the available geographic information system to calculate the distances between RSS and PODs, speed-limit of the roads, etc. The road network was considered as the graph where RSS, POD are the vertices, and roads are edges. The algorithm developed to solve the RSS problem partitions the graph for generating routes. The whole process is illustrated in Figure 1 below:

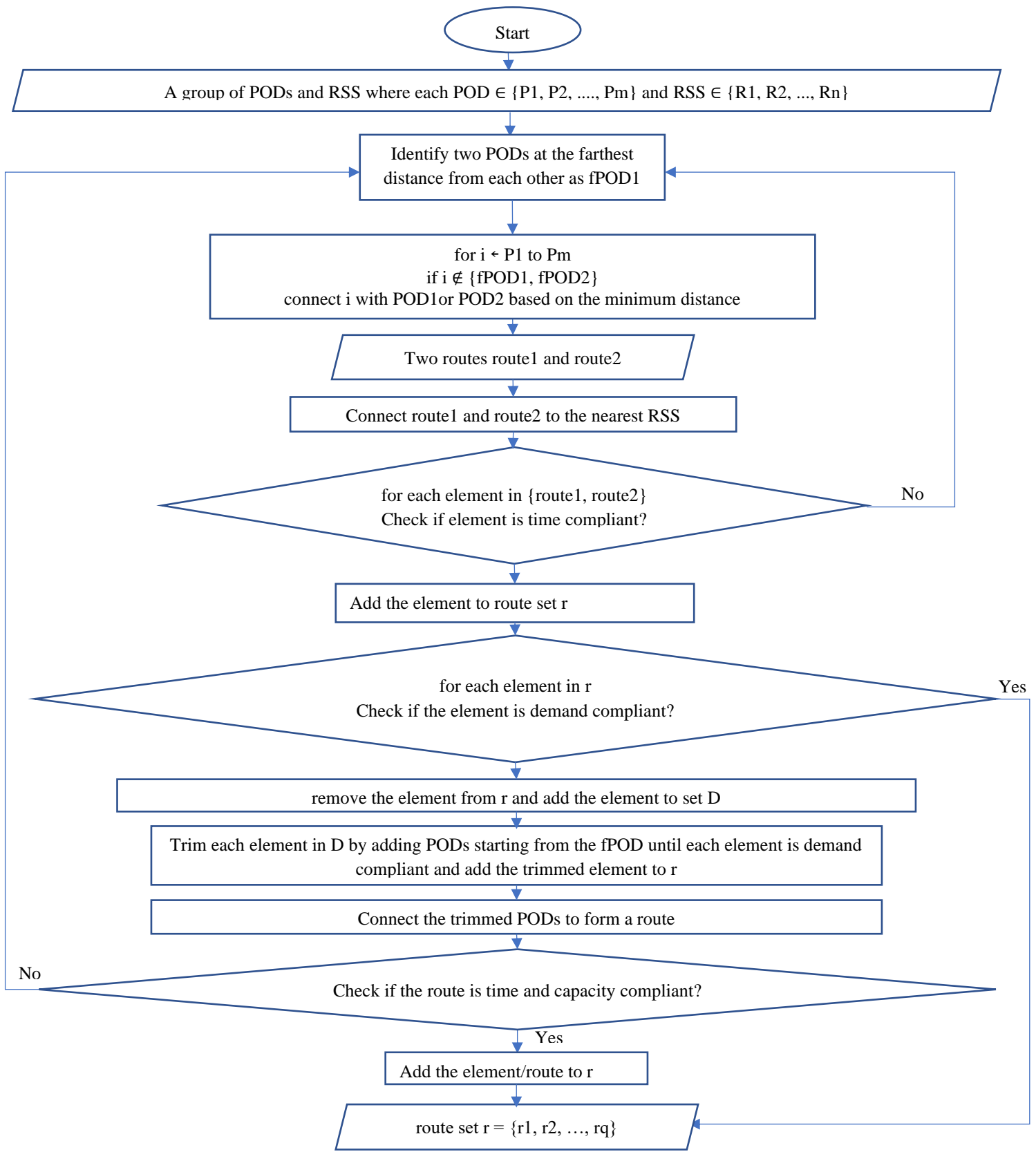

Figure 1: Flowchart of the algorithm to solve the RSS problem.

The route generation process of RSS problem can be explained using the technique used to solve the p-median problem using the following scenario, where we consider both POD and RSS as locations. Two PODS, which were at the farthest distance from each other, were chosen to create two routes. These two PODs are named fPOD1 and fPOD2 for a better understanding of the algorithm. The distance is represented using dist (i, j) where i, j ∈ {POD, fPOD1, fPOD2, RSS}. The other PODs were assigned to either fPOD1 or fPOD2 to form routes in the following way: From both fPOD1 and fPOD2, the closest POD was searched. If the closest POD from fPOD1 is POD3, and the closest POD from fPOD2 is POD7. The connections were made to form routes basing on the distance. If the distance between fPOD1 and POD3 is less than the distance between fPOD2 and POD7, then connection was made between fPOD1 and POD3. Otherwise, connection was made between fPOD2 and POD7. The next closest POD was searched and connected in the same way to form routes. After joining all the PODs to either fPOD1 and fPOD2, two routes were formed. Two RSS, called RSS1 and RSS2, are considered for our illustration. Each route has either fPOD1 or fPOD2 in one end and any other POD in the other end. The POD at the last end of the first route is called ePOD1, and the last end of the second route is called ePOD2 for the benefit of understanding the process. The routes were connected to either RSS1 or RSS2 in the following way basing on distance:

For each route, the distance between starting point to RSS1 dist (fPOD1, RSS1), the starting point to RSS2 dist (fPOD1, RSS2), ending point to RSS1 dist (ePOD1, RSS1), and ending point to RSS2 dist (ePOD2, RSS2) were calculated. The pair which had the least distance was connected. Each route was connected to either RSS1 or RSS2, and the time required to travel the entire route was calculated. Each of the routes was checked if it could deliver within the time limit. The total time required to travel the entire route is represented by $T_t$, and the time limit is represented $T_l$.

Each route was chosen, and for each route, if $T_t > T_l$ then the following steps were taken:
1. the link between RSS and route was disconnected
2. all connections in the route were disconnected
3. the route was partitioned

The process described was continued until all routes are time compliant.

The link between fPOD and POD is represented using $X_{ij}$ where

$X_{ij}$ = 1, if point i is POD and j is fPOD.
   = 0, otherwise.

The link between POD or RSS is represented using $S_{jk}$ where

$S_{jk}$ = 1, if point j is fPOD, and k is POD or RSS.
   = 0, otherwise.

The link between two different PODs is represented using $Y_{ij}$ where

$Y_{ij}$ = 1, if i, j are different PODs and i < j.

= 0, otherwise.

The goal was to minimize for each route,

$$\sum_{i=1}^{n} \sum_{j=1}^{n} (X_{ij} d_{ij} + \sum_{k=1}^{n} S_{jk} Y_{ik} d_{ik})$$

In the p-median problem, it is assumed that the demand is equal for all demand points, and each facility can meet the demand for all demand points. But, in the RSS problem, demands for all the PODs are not the same. Each POD has different demand according to which the vehicles will deliver the SNS assets. The algorithm was designed in a way that the combined demand of all the PODs in a route is less than or equal to the capacity of the vehicle. If the total demand for all the PODs in the route is more than this value, then the vehicle dedicated to each route will not be able to meet the demand of all the PODs in the route. After time compliant routes were formed using the algorithm, the trimming of routes was applied to make it demand compliant. For each route not compliant with demand, the following process was implemented:

PODs were added to the route starting from the POD farthest from RSS, and the process was continued until the combined demand of the PODs crosses the demand limit. After trimmed PODs were found from all the routes, it was connected to form a new route. If the new route was found to be not compliant with time or capacity or both, the whole algorithm was applied again to generate time and capacity compliant route.

**Dataset Description and Analysis:**

Three datasets were used to evaluate the algorithm's performance. The first two describe the test cases fed into the algorithm. These are dataset E collected from [27] and dataset F collected from [26]. Both datasets are lists of coordinates on a grid along with a population, which represents the amount of people who would need to be served. Each testcase also comes with a predefined minimized number of vehicles and a defined capacity for each vehicle. The datasets were sourced from referenced papers and parsed. The first line of each testcase describes the source of the vehicles, and the following 'n-1' lines contain three numbers describing each node – the 'x' coordinate of the node, the 'y' coordinate of the node, and the population of the node. Finally, the last line of the testcase file contains two numbers —the vehicle capacity and the maximum allowed time/cost for the longest route. The original test case file does not include a maximum allowed time/cost for the longest route, but we must add this for reasons discussed later. Each test case also has a code that describes it. After a hyphen, there is the letter 'n' followed by the number of coordinates, which is the number of nodes plus the source. Then, another hyphen, the character 'k', and the predefined minimized number of vehicles of the test case. The reason that each test case comes with a vehicle is because the test cases are to solve a problem different than what our algorithm aims to solve. F and E datasets were used to try to minimize the total cost of all routes given a fixed vehicle capacity and vehicle number. We, however, are trying to minimize the number of vehicles/routes given a fixed vehicle capacity and a maximum time/cost for the longest route.

Example of parsed test case file (E-n22-k4):

| Line 1 | Describes the source/RSS. | 145 215 0 |
| --- | --- | --- |
| Line 2 – n | Describes each node. | 151 264 1100 <br><br>… <br><br>139 182 700 |
| Line n+1 | Vehicle capacity and maximum allowed time/cost for the longest route. | 6000 90 |

The third dataset collected from paper by Lysgaard [28] was used to find the most recent best-case timings and routes for each of the test cases in the F and E datasets. For each test case in the E and F datasets, this dataset provided k lists of integers, where k is the predefined minimized number of vehicles for the original test case. Each list of integers listed node IDs in the order of their occurrence in the ideal route. A node ID is simply the line number of the node's coordinates and population in the test case file. Each best-case data point has the same code as its corresponding test case in the E and F datasets, with the prefix 'opt-'.

Example of solution file (opt-E-n22-k4):

```
Route #1: 17 20 18 15 12
Route #2: 16 19 21 14
Route #3: 13 11 4 3 8 10
Route #4: 9 7 5 2 1 6
Cost: 375
```

Earlier, it was said that the E and F dataset was used to solve a different problem. How, then, do we use its best-case solutions to evaluate the results of our algorithm, which takes in different inputs? Also, why do we have to add a maximum time/cost for the longest route to our test case file? Because each test set comes with a predefined minimized number of routes, we can input each test case's nodes, vehicle capacity, and the largest time/cost of the longest route in the solution of the test case from paper [28]. We can then see how many routes our algorithm generates and compare it to the minimized number of routes for each file to see the effectiveness of our algorithm. Summary of the routes and its associated costs are presented in Table 1 below.

Table 1: Data for Arbitrarily Selected Maximum Route and its associated Costs (time in second)

| Testcase | Max Cost of longest route in solution (time in second) | Input Max Cost of longest route (time in second) | Maximum cost of the longest route generated (time in second) | Actual Minimum Number of Routes | Number of Routes generated by RSS algorithm | Difference= (Number of Routes generated by RSS algorithm)- (Actual Minimum Number of Routes) |
|---|---|---|---|---|---|---|
| E-n22-k4 | 85.317 | 90 | 67.65 | 4 | 5 | 1 |
| E-n23-k3 | 268.085 | 280 | 227.22 | 3 | 6 | 3 |
| E-n30-k3 | 169.947 | 190 | 111.392 | 3 | 5 | 2 |
| E-n33-k4 | 198.661 | 200 | 178.641 | 4 | 6 | 2 |
| E-n51-k5 | 97.952 | 100 | 95.869 | 5 | 8 | 3 |
| E-n76-k7 | 106.038 | 120 | 112.79 | 7 | 9 | 2 |
| F-n45-k4 | 407.807 | 420 | 201.431 | 4 | 5 | 1 |
| F-n72-k4 | 65.972 | 80 | 77.997 | 4 | 7 | 3 |
| F-n135-k7 | 216.088 | 230 | 226.258 | 7 | 9 | 2 |

After generating the routes for each test case here, however, we saw some problems. The routes we generated had costs that were very disparate, and some routes had much higher costs than others. This is illustrated in the Table 2 below.

Table 2: Mean and Range of Generated Route Costs

| Testcase | mean route cost (time in second) | range of route costs (time in second) |
|---|---|---|
| E-n22-k4 | 55.31646235 | 23.74258558 |
| E-n23-k3 | 94.30039942 | 205.015922 |

| Testcase | | |
|---|---|---|
| E-n30-k3 | 89.15826883 | 95.58104541 |
| E-n33-k4 | 112.6450211 | 130.3086072 |
| E-n51-k5 | 81.11862795 | 34.56547722 |
| E-n76-k7 | 80.6198377 | 109.7904589 |
| F-n45-k4 | 133.0773595 | 130.5182428 |
| F-n72-k4 | 42.25389733 | 63.25849456 |
| F-n135-k7 | 115.7127918 | 190.7221814 |

For some of the test cases (E-n22-k4, E-n23-k3, E-n30-k3, and F-n45-k4) it seems that the generated time/cost of the longest route is nowhere near the input maximum time/cost of the longest route. How does this input affect our results? After adjusting the input maximum time/cost of the longest route, the algorithm was able to generate much better results.

Table 3: Comparison of the result from the best case in literature

| Testcase | Input Maximum time/cost of longest route (time in second) | number of routes generated by RSS algorithm | Time/Cost of the Longest Route (time in second) | Difference from Best Case Number of Routes in literature |
|---|---|---|---|---|
| E-n22-k4 | 82.7 | 5 | 84 | 1 |
| E-n23-k3 | 165 | 6 | 163 | 3 |
| E-n30-k3 | 111.4 | 5 | 111.392 | 2 |
| E-n33-k4 | 221 | 5 | 244.625 | 1 |
| E-n51-k5 | 231 | 6 | 125.247 | 1 |
| E-n76-k7 | 199 | 7 | 220.58 | 0 |
| F-n45-k4 | 350 | 5 | 230 | 1 |
| F-n72-k4 | 125 | 6 | 78 | 2 |
| F-n135-k7 | 462 | 8 | 215 | 1 |

**Results**

It is observed from Figure 2 that the technique used was successful in reducing the difference between the number of routes generated by the best case in literature and the number of routes generated by the RSS algorithm for 5 test cases out of 9 test cases.

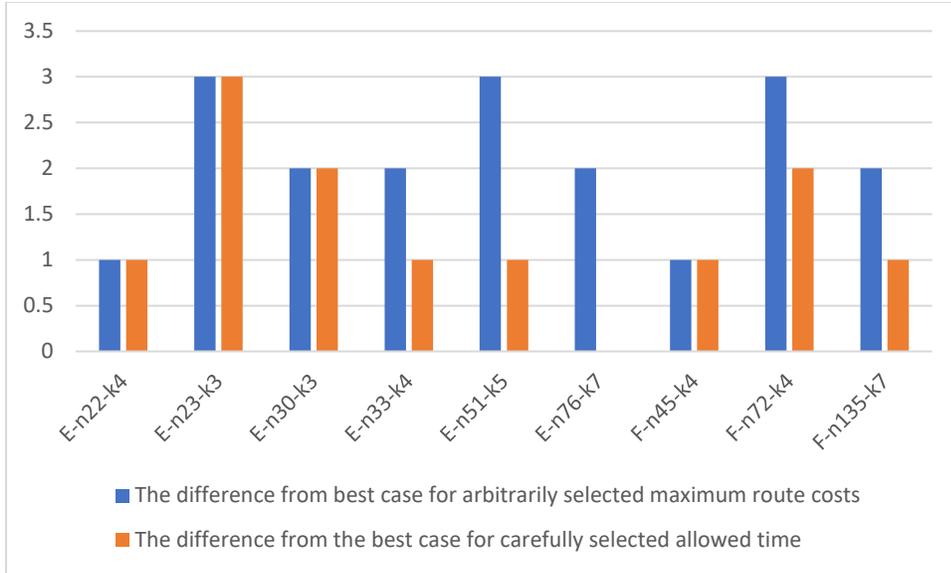

Figure 2: The difference in the number of routes generated by the RSS algorithm and the best case in the literature using two different techniques.

For this reason, data for Carefully Selected Maximum Allowed Time/Costs in Table 3 were chosen to evaluate how good the RSS algorithm performs compared to the best case.

The number of routes generated by the best case in literature and the number of routes generated by the RSS algorithm for each test case is plotted in Figure 3. It is found that the difference in the number of routes generated in each of the test cases for two different algorithms is very close. It is noted that the highest distinction in the number of routes is 3, which is found for test case E-n23-k3. And the lowest difference in the number of routes is 0, which is found for test case E-n76-k7. The RSS algorithm gives the same result as the best case in literature for E-n76-k7. The average difference from the best case is found using the following equation. Average $= \frac{\sum \text{Sum of differnce in number of routes for each testcase}}{\text{Total number of testcase}} = \frac{1+3+2+1+1+0+1+2+1}{9} = \frac{12}{9} = 1.3$

It is found using the above equation that the difference is around 1 on average. It is observed that the RSS algorithm gives a very close result compared to the best case in literature.

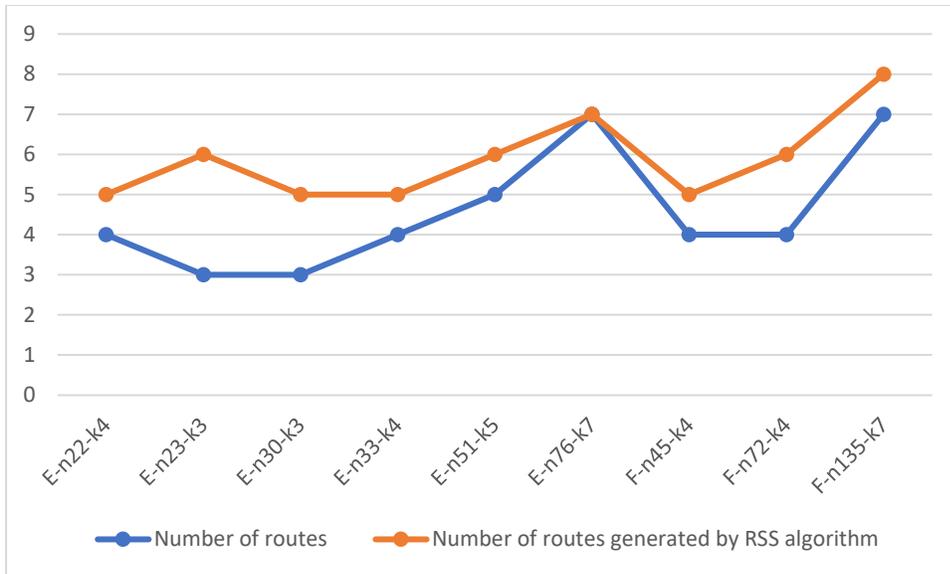

Figure 3: The number of routes generated for best case in literature and the number of routes generated by the RSS algorithm.

**Discussion and conclusion**

There is no way to compare the results of our algorithm with the optimal solution. For this reason, our algorithm was compared with the best case in literature. The method chosen by us for understanding how close our algorithm is compared to the best-case gave was a pretty satisfactory result. It was observed that that the highest cost out of all routes generated by our algorithm was not close to the maximum cost that was specified, which indicates that the RSS algorithm could give us a better result. There are chances that if a better way is found to compare the results of our algorithm with the best case in literature, more satisfactory results can be found. Several tools were analyzed that generate routes for distribution. None of them focuses explicitly on the particularities and challenges that are faced by public health practitioners. Some of the route planning software tools available do not account for stoppage time, and some of them do not minimize waste in transportation by sending trucks that are not fully loaded. It is a big problem in emergency management because of the time-sensitivity in public emergencies. There is no space for waste when a matter of seconds can be translated into saving human lives. That is why our algorithm serves the best interest of the public health planners by tailoring our design to meet their needs. This paper presents a new and innovative algorithm that focuses on the peculiarities of distribution in the context of public health emergency management. It proposes an algorithm that takes into account loading time and stoppage time that is crucial in public health planning. It also considers multiple RSS, which are an essential element in public health planning. It runs in an acceptable time and generates a feasible solution that is required in emergency response planning and poses acceptable computational complexity, which makes it a good alternative for implementation in the public health domain.

**Conflict of interest:** To the best of my knowledge, there is no conflict of interest.

**Data availability and code for RSS algorithm:** Algorithms implemented and datasets used are available in the link below: https://github.com/ArnavIyer/RSS-Algorithm

**Funding:** The whole research was funded by Texas Department of State Health Services.